\documentstyle[aaai]{article} 

\newcommand{\n}{\mbox{$:$}}
\newcommand{\setslash}{\mbox{ $|$ }}
\newcommand{\rev}{{\mbox{$*$}}}
\newcommand{\flies}{{\it flies}}
\newcommand{\default}{{\it default}}
\newcommand{\lfp}{{\it lfp}}

\newenvironment{proof}{{\bf Proof\mbox{:\hspace{.02in}} }}{\( \Box \) \vspace{.1in}}

\newtheorem{definition}{Definition}
\newtheorem{proposition} {Proposition}

\title{Declarative Representation of Revision Strategies}
\author{Gerhard Brewka \\
University of Leipzig, Dep. of Computer Science \\
Augustusplatz 10-11, 04109 Leipzig, Germany \\
brewka@informatik.uni-leipzig.de}

\begin{document}

\maketitle

\begin{abstract}
In this paper we introduce a nonmonotonic framework for belief revision in which reasoning about the reliability of different pieces of information based on meta-knowledge about the information is possible, and where revision strategies can be described declaratively. The approach is based on a Poole-style system for default reasoning in which entrenchment information is represented in the logical language. A notion of inference based on the least fixed point of a monotone operator is used to make sure that all theories possess a consistent set of conclusions.

\end{abstract}

\section{Introduction}
Formal models of belief revision differ in what they consider as representations of the epistemic states of an agent. In the AGM approach \cite{Gar88} epistemic states are identified with logical theories, that is, sets of formulas closed under classical inference. Other approaches like those discussed in \cite{Neb92} consider finite sets of formulas, sometimes called belief bases, as epistemic states. They investigate how to revise such belief bases. Only a rather small fraction of work in belief revision has studied an obvious alternative: the revision of epistemic states expressed as nonmonotonic theories \cite{Bre91,WilAnt98,Antetal99,ChoPar99}.

This is somewhat surprising since close relationships between properties of nonmonotonic inference relations and postulates for belief revision have been established \cite{MakGar91,GarMak94}. Indeed, one of the reasons why nonmonotonic logics were invented is their ability to handle conflicts and inconsistencies, one of the major issues in belief revision. If this is the case, shouldn't it be possible to use the power of nonmonotonic inference to simplify revision? In fact, what we have in mind is a complete trivialization of the revision problem. We want to be able to revise nonmonotonic theories simply by adding new information, and we want to leave everything else to the nonmonotonic inference relation. 

An early approach in this spirit was the author's paper \cite{Bre91} where an extension of Poole-systems \cite{Poo88}, the so-called preferred subtheory approach, was used. New information, possibly equipped with information about the reliability level of this information, was simply added to the available information. The nonmonotonic inference relation determined the acceptable beliefs.

We are not satisfied with this approach any longer for several reasons. Existing theories of belief revision, including the one presented in the earlier paper, have difficulties to model the way real agents revise their beliefs. One of the reasons for this is that they do not represent information which is commonly used by agents for this purpose. For instance, new information always comes together with certain meta-information (formulas don't fly into the agent's mind): Where does the information come from? Was it an observation? Did you read it in the newspaper?  Did someone tell you, and if so, who? Did the person who gave you the information have a motive to lie? and so on. In most cases we reason with and about this meta-information when revising our beliefs. We strongly believe that realistic models of revision should provide the necessary means to represent this kind of information.

The meta-information is used to determine the entrenchment of pieces of information. The less entrenched the information is, the more willing we are to give it up. Again, entrechment relations are not just there, they result from reasoning processes. To model this kind of reasoning, entrenchment should be expressible in the logical language. Once we have the possibility to express entrenchment (or plausibility, or preference) in the language, it will also become possible to represent revision strategies declaratively. This in turn makes it possible to revise the revision strategies themselves.

Here is a real life example that can be used to illustrate what we have in mind. Assume Peter tells you that your girl-friend Anne went out for dinner with another man yesterday. Peter even knows his name: it was John, a highly attractive person known for having numerous affairs. You are concerned and talk to Anne about this. She tells you she was at home yesterday evening waiting for you to call. Peter insists that he saw Anne with that man. You are not sure what to believe. Luckily, you find out that Anne has a twin sister Mary. Mary indeed went out with her new boy-friend John. This explains why Peter got mixed up. You now believe Anne and happily continue your relationship.

What this example nicely illustrates is the way we reason about the reliability of information. There is no given fixed entrenchment ordering to start with. In the example there is also, at least in the beginning, no reason to trust Peter more than the girl-friend, or vice versa. And obviously, it is not the new information that is accepted in each situation. It is the additional context information which is relevant here: it gives us an explanation for Peter's mistake and decreases the reliability of Peter's observation enough to break the tie.

To be able to formalize examples of this kind we propose in this paper an approach to belief revision where
\begin{itemize}
\item nonmonotonic belief bases represent epistemic states and nonmonotonic inference is used to completely trivialize revision,
\item it is possible to express and reason about meta-information, including the reliability of formulas,
\item revision strategies can be represented declaratively, that is, logical formulas express how conflicts among different pieces of information are resolved.
\end{itemize}

The outline of the paper is as follows. In the next section we introduce the nonmonotonic formalism we use here to represent epistemic states. In the following section we show how to use this formalism for representing revision strategies. We then discuss the AGM postulates for revision and show that almost all of them are not valid in our approach (which does not bother us). In the following section we briefly deal with contraction. We then discuss forgetting in the context of our approach. Finally, we discuss related work and conclude.

\section{Representing reliability relations}
In this section we introduce the formalism used in this paper. As mentioned in the introduction, one of the distinguishing features of our approach is that we want to be able to reason about the reliability of the available information in the logical language. In the AGM approach \cite{Gar88,GarMak88} entrenchment relations are used to represent how strongly an agent sticks to his beliefs: the more entrenched a formula, the less willing to give it up the agent is. Entrenchment relations have several properties which are based on the logical strength of the formulas. For instance, logically weaker formulas are not less entrenched than logically stronger ones. The intuition is that if a weaker formula has to be given up, the stronger formula has to be given up anyway. 

In our approach we do not require such properties. We may even have equivalent formulas $p$ and $p'$ with different reliability. This may happen when, for instance, $p$ and $p'$ come from different sources $s$ and $s'$ with different reliability. Note that although the less reliable information does not add to the accepted beliefs as long as the more reliable equivalent information is in force, the situation may change when new information about the reliability of $s$ is obtained. Should $s$ turn out to be highly unreliable later (of course, beliefs about the reliability of sources may be revised as any other beliefs) then it becomes important to have $p'$ with, say, somewhat lower reliability available.

All we require, therefore, is the existence of a strict partial order $\prec$ between formulas which tells us how to resolve potential conflicts. To avoid misunderstandings we will not call $\prec$ an entrenchment relation. Instead, we speak of reliability, or simply priority among formulas.

Since we want to represent $\prec$ in the logical language we need to be able to refer to formulas. Instead of using a quoting mechanism for this purpose, we will use named formulas, that is pairs consisting of a formula and a name for the formula. Technically, names are just ground terms that can be used everywhere in the language.

We will present our formalism in two steps: we first introduce an extension of Poole systems which allows us to express preference information in the language, together with an appropriate definition of extensions. It turns out that due to the potential self-referentiality of preference information not all theories expressed in this formalism possess extensions, that is, acceptable sets of beliefs. In a second step, we therefore introduce a new notion of prioritized inference defined as the least fixed point of a monotone operator. Epistemic states, then, are identified with preferential default theories under this least fixed point semantics.

Our basic formalism extends the well-known Poole systems \cite{Poo88}. Recall that Poole systems consist of a consistent set of (first order) formulas $F$, the facts, and a possibly inconsistent set of formulas $D$, the defaults. A set of formulas $E$ is an extension of a Poole-system $(F,D)$ iff $E = Th(F \cup D')$ where $D'$ is a maximal $F$-consistent subset of $D$.

Our formalism differs from this approach in the following respects:
\begin{enumerate}
\item In the context of belief revision it seems inappropriate to consider some information as absolutely certain and unrevisable. We therefore do not use $F$. Instead, we have a single set $T$ containing all the information.\footnote{One of the reviewers of this paper points out that using $F$ may have representational advantages since it eliminates the need to use preferences to indicate the most reliable information, see our examples in the rest of the paper. We might therefore reintroduce $F$ in future versions of this paper for purely practical reasons. This does not seem to pose any technical problems.}
\item We represent preference and other meta-information in the language. We therefore introduce names for formulas and a special symbol $<$. $d < d'$ intuitively says that in case of a conflict $d'$ should be given up rather than $d$ since the latter is more reliable. We require that $<$ represents a strict partial order.\footnote{We assume that the properties of $<$, like those of equality, are part of the underlying logic and need not be represented through explicit axioms in our default theories.}
\item We introduce a new notion of extension which takes the preference information into account adequately.
\end{enumerate}
To avoid confusion we want to emphasize that $<$ belongs to the logical language, whereas $\prec$ is a meta level symbol. For the following definitions it is essential to clearly separate between these levels.  

For simplicity, we only consider finite default theories in this paper. A generalization to the infinite case would have to reduce partial orderings to well-orderings rather than total orders.
\begin{definition}
A named formula is a structure of the form $d \n p$, where $p$ is a first order formula and $d$ a ground term representing the name of the formula.
\end{definition}
We use the functions $name$ and $form$ to extract the name respectively formula of a named formula, that is $name(d \n p)=d$ and $form(d \n p)=p$. We will also apply both functions to sets of named formulas with the obvious meaning.
\begin{definition}
A preference default theory $T$ is a finite set of named formulas such that
\begin{itemize}
\item $form(T)$ is a set of first order formulas whose logical language contains a reserved symbol $<$ representing a strict total order, and 
\item  $d_1 \n p \in T$, $d_2 \n q \in T$ and $p \not = q$ implies $d_1 \neq d_2$.
\end{itemize}
\end{definition}
The last item in the definition guarantees that different formulas have different names. 
\begin{definition}
Let $T$ be a preference default theory, $\prec$ a total order on $T$. The extension of $T$ generated by $\prec$, denoted $E^{\prec}_T$, is the set $E^{\prec}_T = Th(\bigcup_{i=0}^{|T|} E_i)$ where 
\begin{itemize}
\item $E_0 = \emptyset$, and for $0 < i \leq |T|$
\item $E_{i} = E_{i-1} \cup \{form(d_{i})\}$ if this set is consistent,  $E_{i-1}$ otherwise. \\
Here $d_{i}$ is the $i$-th element of $T$ according to the total order $\prec$. 
\end{itemize}
The set $\bigcup_{i=0}^{|T|} E_i$ is called the extension base of $E^{\prec}_T$.
\end{definition}
We say $E$ is an extension of $T$ if there is some total order $\prec$ such that $E=E^{\prec}_T$. Obviously, all maximal consistent subsets of $form(T)$ are extension bases. We now consider the general case of partial orders.
\begin{definition}
Let $T$ be a preference default theory, $\prec$ a strict partial order on $T$. The set of extensions of $T$ generated by $\prec$ is  \[Ext^{\prec}_T = \{ E^{\prec'}_T \setslash \prec' \mbox{ is a total order extending } \prec\}.\]
\end{definition}
We next define two notions of compatibility:
\begin{definition}
Let $T$ be a preference default theory, $\prec$ a strict partial ordering of $T$, $S$ a set of formulas. We say $\prec$ is compatible with $S$ iff 
\[S \cup \{d < d' \setslash d \n p \prec d' \n q\} \cup  \{\neg(d < d') \setslash d \n p \not \prec d' \n q\}\]
is consistent. \\
An extension $E$ of $T$ is compatible with $S$ iff there is a strict partial ordering $\prec$ of $T$ compatible with $S$ such that $E \in Ext^{\prec}_T$. 
\end{definition}
The set of extensions of $T$ compatible with $S$ is denoted $Ext^S_T$.
\begin{definition}
Let $T$ be a preference default theory. A set of formulas $E$ is called a preferred extension of $T$ iff $E \in Ext^E_T$.
\end{definition}
Intuitively, $E$ is a preferred extension if it is the deductive closure of a maximal consistent subset of $T$ which can be generated through a total preference ordering compatible with the preference information in $E$ itself. The preference information in $E$ certainly does not have to be total.

Here is a simple example illustrating preference default theories:
\begin{quote}
$d_1(x): bird(x) \rightarrow \flies(x) \mbox{    $|$ $x$ ground object term}$ \\
$d_2: \forall x. penguin(x) \rightarrow \neg \flies(x)$ \\
$d_3: bird(tweety) \land penguin(tweety)$ \\
$d_4: \forall x. d_3 < d_1(x)$ \\
$d_5: \forall x. d_2 < d_1(x)$ 
\end{quote}
As is common in Poole systems, rules with exceptions, that is, formulas whose instances can be defeated without defeating the formula as a whole (here $d_1$), are represented as schemata used as abbreviations for all of their ground instances.
As above we will make the intended instances explicit in all examples.
 To make sure that the different ground instances can be distinguished by name we have to parameterize the names also. We assume that terms used as names can be distinguished from other terms which we call object terms.\footnote{A more elaborate formalization would be based on sorted logic with sorts for names and other types of objects from the beginning. We do not pursue this here since we want to keep things as simple as possible.} In our case, $d_1(tweety)$ is a proper rule name, $d_1(d_1)$ is not. Since we only consider finite theories we must also assume that the set of object terms is finite. 

In our example we obtain 3 extensions $E_1$, $E_2$ and $E_3$. In $E_1$ the instance of $d_1(x)$ with $x = tweety$ is rejected, in $E_2$ $d_2$ is rejected, and $E_3$ rejects $d_3$. All extensions contain $d_4$ and $d_5$. It is not difficult to see that only $E_1$ can be constructed using a total ordering of $T$ which is compatible with this information. $E_1$ is thus the single preferred extension of this preference default theory.

Preference default theories under extension semantics are very flexible and expressive. The reason we are not yet fully satisfied with them is that they can express unsatisfiable preference information: there are theories which do not possess any preferred extensions. The simplest example is as follows:
\begin{quote}
$d_1: d_2 < d_1$ \\
$d_2: d_1 < d_2$
\end{quote}
Accepting the first of the two contradictory formulas requires to give preference to the second, and vice versa. No preferred extension exists for this theory.

This means that preference default theories together with the standard notion of nonmonotonic inference where a formula is considered derivable whenever it is contained in all (preferred) extensions do not seem fully adequate for representing epistemic states of rational agents. 

We will therefore introduce another, somewhat less standard notion of nonmonotonic consequence.\footnote{An alternative way of handling this problem would be to introduce some kind of ``stratification'' into our theories. Stratification, a term taken form the area of logic programming, would prohibit formulas from speaking, directly or indirectly, about their own priority. Unfortunately, it turns out that only highly restrictive forms of stratification guarantee existence of extensions. For this reason we do not pursue this approach here.} 
This approach shares some intuition with the fixed point formulation of well-founded semantics for logic programs with negation due to Baral and Subrahmanian \cite{BarSub91}. In particular, it is based on the least fixed point of a monotone operator. 

Let us first explain the underlying idea.  Starting with the empty set, we iteratively compute the intersection of those extensions which are compatible with the information obtained so far. Since the set of formulas computed in each step may contain new preference information the number of extensions may be reduced, and their intersection thus may grow. We continue like this until no further change happens, that is, until a fixed point is reached.
\begin{definition} \label{def:Cp}
Let $T$ be a preference default theory, $S$ a set of formulas. We define an operator $C_T$ as follows:
\[C_T(S) = \bigcap Ext^S_T \]
\end{definition}
\begin{proposition}
The operator $C_T$ is monotone.
\end{proposition}
\begin{proof}
$S \subseteq S'$ implies that an ordering $\prec$ is compatible with $S$ whenever it is compatible with $S'$. We thus have $Ext^{S'}_T \subseteq Ext^{S}_T$ and therefore $ \bigcap Ext^S_T \subseteq  \bigcap Ext^{S'}_T$. 
\end{proof}

Monotone operators, according to the well-known Knaster-Tarski theorem \cite{Tar55}, possess a least fixed point. This fixed point can be computed by iterating the operator on the empty set. We, therefore, can define the accepted conclusions of a preference default theory as follows:
\begin{definition}
Let $T$ be a preference default theory. A formula $p$ is an accepted conclusion of $T$ iff $p \in \lfp(C_T)$, where $\lfp(C_T)$ is the least fixed point of the operator $C_T$.
\end{definition}
We call extensions which are compatible with $\lfp(C_T)$ accepted extensions.

Several illustrative examples will be given in the next section. 
Here we just show how the theory without preferred extension is handled in this approach.
We have $T = \{d_1 \n  (d_2 < d_1), d_2 \n (d_1 < d_2)\}$. We first compute
$C_T(\emptyset)$. Since no preference information is available in the empty
set we obtain $Th(\{d_2 < d_1\}) \cap Th(\{d_1 < d_2\})$ which is equivalent to
$Th(\{d_2 < d_1 \vee d_1 < d_2\})$. This set is already the least fixed point.

\begin{proposition} Let $T$ be a preference default theory, $p$ an
accepted conclusion of $T$. Then $p$ is contained in all preferred
extensions of $T$.
 \end{proposition} 
\begin{proof}
If $T$ has no preferred extension the proposition is trivially true. So assume $T$ possesses preferred extension(s). A simple induction shows that each preferred extension is among the extensions compatible with the formulas computed in each step of the iteration of $C_T$. Therefore each preferred extension is also an accepted extension.
\end{proof}

\begin{proposition} \label{prop:cons}
Let $T$ be a preference default theory. The set of accepted conclusions of $T$ is consistent.
\end{proposition}
\begin{proof}
We show by induction that, for arbitrary $n$, the set of formulas obtained after $n$ applications of $C_T$ is consistent. For $n = 0$ this is trivial. Assume the set of formulas $S$ obtained after $n-1$ iterations is consistent. Since $S$ is consistent and $<$ formalizes a strict partial ordering there must be at least one strict partial ordering $\prec$ compatible with $S$, so the set of all extensions compatible with $S$ is nonempty. Since each extension is by definition consistent the intersection of an arbitrary nonempty set of extensions must also be consistent.
\end{proof}

Since preference default theories under accepted conclusion semantics always lead to consistent beliefs, we will in the next section identify epistemic states with preference default theories and belief sets with their accepted conclusions.

\section{Revising epistemic states}
\subsection{The revision operator}
Given an agent's epistemic state is identified with a preference default theory as introduced in the last section, it is natural to identify the set of beliefs accepted by the agent with the accepted conclusions of this theory. We therefore define belief sets as follows:
\begin{definition}
Let $T$ be an epistemic state. $Bel(T)$, the belief set induced by $T$, is the set of accepted conclusions of $T$.
\end{definition}
It is a basic assumption of our approach that belief sets cannot be revised directly. Revision of belief sets is always indirect, through the revision of the epistemic state inducing the belief set. Note that since two different epistemic states may induce the same belief set, the revision function which takes an epistemic state and a formula and produces a new epistemic state does not induce a corresponding function on belief sets.

Given an epistemic state $T$, revising it with new information simply means  generating a new name for it and adding the corresponding named formula.
\begin{definition}
Let $T$ be an epistemic state, $p$ a formula. The revision of $T$ with $p$, denoted $T \rev p$, is the epistemic state $(T \cup \{n \n p\})$ where $n$ is a new name not appearing in $T$. 
\end{definition}
Notation: in the rest of the paper we assume that names are of the form $d_j$ where $j$ is a numbering of the formulas. If $T$ has $j$ elements and a new formula is added, then its new name is $d_{j+1}$.

\subsection{Representing revision strategies} \label{sec:examples}
In this subsection we show how revision strategies used by an agent can be represented in our approach. We first discuss an example where the strategy is based on the type of the available information. We distinguish between strict rules, observations and defaults. Strict rules have highest priority because they represent well-established or terminological information. Observations can be wrong, but they are considered more reliable than default information. Consider the following epistemic state $T$:
\begin{quote}
$d_1: penguin(tweety)$\\
$d_2:\forall x. penguin(x) \rightarrow bird(x)$\\
$d_3: \forall x. penguin(x) \rightarrow \neg \flies(x)$\\
$d_4(x):bird(x) \rightarrow \flies(x) \mbox{    $|$ $x$ ground object term}$\\
$d_5:observation(d_1)$\\
$d_6:rule(d_2)$\\
$d_7:rule(d_3)$\\
$d_8:\forall x. \default(d_4(x))$\\
$d_9:\forall n,n'. rule(n) \land observation(n') \rightarrow n < n'$\\
$d_{10}:\forall n,n'. observation(n) \land default(n') \rightarrow n < n'$
\end{quote}
$T$ has 4 extensions. The corresponding extension bases are obtained from $T$ by leaving out 
$d_1$, $d_2$, $d_3$, or $d_4(tweety)$, respectively. All extensions, and thus $C_T(\emptyset)$, contain information stating that $d_4(tweety)$ has lower preference than the other three formulas. Therefore, the only extension compatible with $C_T(\emptyset)$ is the one generated by leaving out $d_4(tweety)$. This set is also the least fixpoint of $C_T$.
$Bel(T)$ thus does not contain $\flies(tweety)$. 

The next example formalizes the revision strategy of an agent who prefers newer information over older information and information from a more reliable source over information from a less reliable source. In case of a conflict between the two criteria the latter one wins. 

Assume the following specific scenario: At time 10 Peter informs you that $p$ holds. At time 11 John tells you this is not true. Although you normally prefer later information, you also have reason to prefer what Peter told you since you believe Peter is more reliable than John. Since you consider reliability of your sources even more important than the temporal order you believe $p$.

Here is the formal representation of this scenario. We use $X < d$ where $X$ is a finite set of names as an abbreviation for $\bigwedge_{x \in X} x < d$. Note that we have to make sure, by adding adequate preferences, that the rules representing our revision strategy cannot be used - via contraposition - to defeat our meta-knowldge about $d_1$ and $d_2$:

\begin{tabbing}
X \= XXXXXX \= XXXXXXXXXXX \= \kill
\> $d_1: p$\\
\> $d_2: \neg p$\\
\> $d_3:time(d_1) = 10$\\
\> $d_4:time(d_2) = 11$\\
\> $d_5:source(d_1) = Peter$\\
\> $d_6:source(d_2) = John$\\
\> $d_7:more \mbox{-} rel(John,Peter)$\\
\> $d_8(n,n'):more \mbox{-} rel(source(n),source(n')) \rightarrow  n < n'$ \\
\> \> \> $ \mbox{    $|$ $n,n' \in \{d_1,\ldots, d_7\}$}$\\
\> $d_9(n,n'):time(n) < time(n') \rightarrow n' < n $ \\
\> \> \> $\mbox{    $|$ $n,n' \in \{d_1,\ldots, d_7\}$}$\\
\> $d_{10}:\forall n, n'. \;  \{d_3, \ldots, d_7\} < d_8(n,n') < d_9(n,n')$
\end{tabbing}

This preference default theory has 14 extensions which are obtained by leaving out one of $\{d_1,d_2\}$ and one of $\{d_3, d_4, d_5, d_6, d_7, d_8(d_1,d_2), d_9(d_1,d_2)\}$. All extensions contain $d_{10}$. This means that after the next iteration of the $C_T$-operator we are left with 2 extensions which are obtained by leaving out one of $\{d_1,d_2\}$ and $d_9(d_1,d_2)$. Both extensions contain the formula $d_1 < d_2$. The next and final iteration of $C_T$ thus eliminates the extension containing $d_2$. We are left with a single extension and $p$ is among the accepted conclusions.

We next present the example from the introduction. This time we use categories $low$, $medium$ and $high$\footnote{We assume uniqueness of names for the categories. Otherwise the set $\{d_3, d_5, d_6, d_7\}$ would be consistent and could be used to defeat $d_9$ which, obviously, is unintended.} to express reliability: the reliability of a formula with name $n$ is $rel(n)$.  We have the following information:

\begin{tabbing}
X \= XXX \= XXXXXXXXXXXX \= \kill
$d_1: date(Anne,John)$\\
$d_2: \neg date(Anne,John)$\\
$d_3: rel(d_1) = medium$\\
$d_4:rel(d_2) = medium$\\
$d_5: date(Mary,John)$\\
$d_6:twins(Mary, Anne)$\\
$d_7: date(Mary, John) \land twins(Mary, Anne) \rightarrow$ \\
\> \>  $rel(d_1) = low$\\
$d_8:\forall n, n'. \; rel(n) = high \land rel(n') = medium \rightarrow $ \\
\> \>  $n < n'$\\
$d_9:\forall n, n'. \; rel(n) = medium \land rel(n') = low \rightarrow $ \\
\> \>  $ n < n'$\\
$d_{10}: rel(d_5) = rel(d_6) = rel(d_7) = rel(d_8) = $ \\
\> \>  $rel(d_9) = high$\\
$d_{11}:rel(d_3) = rel(d_4) = medium$ 
\end{tabbing}
Although the agent initially considers $d_1$ and $d_2$ as equally reliable, the information that Anne has a twin sister Mary who is dating John decreases the reliability of $d_1$ to $low$. $d_8$ and $d_9$ say how the reliability categories are to be translated to preferences.  $d_{10}$ and $d_{11}$ make sure that meta-information is preferred, and that $d_7$ can defeat $d_3$.Taking all reliability information into account the agent accepts $\neg date(Anne,John)$.

Note that we do not discuss here how agents form meta-beliefs of the kind required to represent the example (induction, folk psychology?). We simply assume that this information is available to the agent. 

\section{Postulates}
We now discuss the postulates for revision which are at the heart of the AGM approach \cite{Gar88}. Since our approach uses epistemic states rather than deductively closed sets of formulas (belief sets) as substrate of revision, some of the postulates need reformulation. In particular, AGM use the expansion operator $+$ in some postulates. Expansion of a belief set $K$ with a formula $p$ means adding $p$ to the belief set and closing under deduction, that is $K+p = Th(K \cup \{p\})$. Since epistemic states always induce consistent belief sets the distinction between revising and expanding an epistemic state does not seem to make much sense in our context. We therefore translate expansion in the following postulates to expansion of the induced belief set. 

In the following we present the AGM postulates (K*i) in their original form together with our corresponding reformulations (T*i). In each case K is a belief set in the sense of AGM, $T$ an epistemic state as defined in this paper, $A, B$ are formulas:
\begin{quote}
(K*1) $K \rev A$ is a belief set. \\
(T*1) $Bel(T \rev A)$ is belief set.
\end{quote}
Obviously satisfied.
\begin{quote}
(K*2) $A \in K \rev A$ \\
(T*2) $A \in Bel(T \rev A)$ 
\end{quote}
Not satisfied. New information is not necessarily accepted in our approach. We see this as an advantage since otherwise belief sets would always depend on the order in which information was obtained.
\begin{quote}
(K*3) $K \rev A \subseteq K+A$ \\
(T*3) $Bel(T \rev A) \subseteq Bel(T)+A$
\end{quote}
Not satisfied. Assume we have $T = \{d_1 \n p, d_2 \n  \neg p \}$, that is $Bel(T)$ is the set of tautologies. Let $A = d_1 < d_2$. Now $Bel(T \rev A)$ contains $p$ which is not contained in $Bel(T)+A$.
\begin{tabbing}
xxxxi \= \kill
(K*4) if $\neg A \not \in K$ then $K+A \subseteq K \rev A$ \\
(T*4) if $\neg A \not \in Bel(T)$ then \\
\> $Bel(T)+A \subseteq Bel(T \rev A)$ 
\end{tabbing}
Not satisfied. It may be the case that $\neg A$, although not in the belief set, is contained in one of the accepted extensions. Adding $A$ to the epistemic state does not necessarily lead to a situation where this extension disappears.
\begin{quote}
(K*5) $K \rev A \vdash \bot$ iff $\vdash \neg A$ \\
(T*5) $Bel(T \rev A) \vdash \bot$ iff $\vdash \neg A$
\end{quote}
Not satisfied. Revising an epistemic state with logically inconsistent information has no effect whatsoever. The information is simply disregarded. Inconsistent belief sets are impossible in our approach, so the right to left implication does not hold.
\begin{quote}
(K*6) If $A \leftrightarrow B$ then $K \rev A = K \rev B$ \\
(T*6) If $A \leftrightarrow B$ then $Bel(T \rev A) = Bel(T \rev B)$ 
\end{quote}
Satisfied under the condition that $A$ and $B$ are given the same name, or the names of $A$ and $B$ do not yet appear in $S$. But note that logically equivalent information may have different impact on the belief sets when different meta-information is available. For instance, $d_1 \n p$ and $d_2 \n p$ may have different effects if different meta-information about the sources of $d_1$ and $d_2$, respectively, is available.
\begin{quote}
(K*7) $K \rev (A \land B) \subseteq (K \rev A)+B$ \\
(T*7) $Bel(T \rev (A \land B)) \subseteq Bel(T \rev A)+B$ 
\end{quote}
Not satisfied. Here is a counterexample. Assume we have $T = \{d_1 \n p, d_2 \n  \neg p, d_3 \n  \neg p\}$. Now let $A = d_1 < d_2$ and $B = d_1 < d_3$. Clearly, revising the epistemic state with $A \land B$ leads to a single accepted extension containing $p$ since the two conflicting formulas are less preferred. $p$ is thus in the belief set induced by the revised state. On the other hand, revising the epistemic state with $A$ leads to two extensions, one containing $p$, the other $\neg p$. $p$ is thus not in the belief set induced by the new state. This does not change when we expand the belief set with $d_1 < d_3$. 
\begin{tabbing}
xxxxi \= \kill
(K*8) If $\neg B \not \in K \rev A$ then  $(K \rev A)+B \subseteq   K \rev (A \land B)$\\
(T*8) If $\neg B \not \in Bel(T \rev A)$ then  \\
\> $Bel(T \rev A)+B \subseteq   Bel(T \rev (A \land B))$
\end{tabbing}
Not satisfied. This is immediate from the fact that $Bel(T \rev (A \land B))$ does not necessarily contain $B$, that is from the failure of (T*2).

This analysis shows that the intuitions captured by the AGM postulates are indeed very different from those underlying our approach. 

\section{Contraction}
Contraction means making a formula underivable without assuming its negation. There may be different reasons for this, not all of them requiring extensions of our framework. For instance, the reliability of a source of a certain piece of information may be in doubt due to extra information. In that case it may happen that a belief $p \in Bel(T)$ is no longer in the belief set $Bel(T \rev q)$ for appropriate $q$ even if $\neg p \not \in Bel(T \rev q)$. Such effects are handled implicitly in our approach.

If, however, the agent may obtain information of the kind ``do not believe $p$'' rather than ``believe $\neg p$'', then extra mechanisms seem necessary. In the context of AGM-style approaches the contraction operator $-$ can be defined through revision on the basis of the so-called Harper identity: $K-A = (K \rev \neg A) \cap K$. The intuition here is that revision with $\neg A$ removes the formulas used to derive $A$, and the intersection with $K$ guarantees that no new information is derived from $\neg A$. 

This intuition can, to a certain extent, be captured using Poole's constraints \cite{Poo88}. Constraints, basically, are formulas used in the construction of maximal consistent subsets of the premises, but not used for derivations. 

To model contraction of epistemic states we must distinguish between these two types of formulas, premises and constraints. Extension bases consist of both types and also the compatibility of preference orderings is checked against premises and contraints. Extensions, however, are generated only from the premises. Constraints, as regular formulas, have names and may come with meta-information, e.g., information about their reliability.

We do not want to go into further technical detail here. Instead, we illustrate contraction using an example. We indicate constraints by choosing names of the form $c_j$ for them. Assume the epistemic state is as follows:
\begin{quote}
$d_1: peng(tweety)$\\
$d_2(x):peng(x) \rightarrow \neg \flies(x)$
\end{quote}
The agent receives the information ``do not believe $\neg flies(tweety)$''. The following constraint is added:
\begin{quote}
$c_1: \flies(tweety)$
\end{quote}
Note that the constraint is not necessarily preferred to the premises. Let $inst(d_2)$ denote the set of all ground instances of $d_2$. We obtain three extension bases 
\begin{quote}
$E_1 = \{peng(tweety)\} \cup inst(d_2)$ \\
$E_2 = \{peng(tweety), \flies(tweety)\} \, \cup \, inst(d_2) \setminus \{peng(tweety) \rightarrow \neg \flies(tweety) \}$ \\
$E_3 = \{\flies(tweety)\} \cup inst(d_2)$ 
\end{quote}
Although $E_2$ and $E_3$ contain $\flies(tweety)$ this formula is not in the extensions generated from these extension bases, and for this reason not in the belief set, since it is a constraint. 

Note that constraints do not necessarily prohibit formulas from being in the belief set since they may have low reliability. For example, if we revise the epistemic state obtained above with $d_1 < c_1$ and $\forall x. d_2(x) < c_1$ then the belief set contains $\neg \flies(tweety)$.

Although we used the Harper identity above to motivate the use of constraints for contraction, its natural reformulation 
\[Bel(T-A) = Bel(T \rev \neg A) \cap Bel(T)  \]
is not valid in our approach.  Assume $T = \{d_1 \n p, d_2 \n  \neg p\}$. Obviously, $Bel(T)$ is the set of tautologies. Now let $A = \neg (d_1 < d_2)$. We contract by adding the constraint $c_1 \n  (d_1 < d_2)$. Now the single accepted extension of the new epistemic state and thus its belief set is $Th(p)$, a strict superset of $Bel(T)$.

\section{Forgetting}
In our approach revising a knowledge base means adding a formula to the epistemic state. Even in the case of contraction the epistemic state grows. For ideal agents this may be adequate since every piece of information, whether it contributes to the current belief set or not, may turn out to be relevant later. However, for agents with limited resources the expansion of the epistemic state cannot go on forever. 

This raises the question how and when pieces of information should be forgotten. What we need is some kind of a mental garbage collection strategy. In LISP systems garbage collection is the process of identifying unaccessable memory space which is then made available again. In our context there is no clear distinction between garbage and non-garbage. As mentioned before, every information may become relevant through additional information, so it would not be reasonable to throw away information just because it is, say, not contained in any extension base. On the other hand, even those formulas contributing to the current belief set may be considered as garbage if the corresponding part of the belief set is not relevant to the agent. It appears that a satisfactory treatment of forgetting would have to take the utility of information for the agent into account. This is beyond the scope of this paper and a topic of further research.

\section{Related work and discussion}
In this paper we proposed a framework for belief revision where preference default theories together with a corresponding nonmonotonic inference relation are used to represent epistemic states and belief sets, respectively. Our underlying formalism draws upon ideas developed in \cite{Bre89} and \cite{BreEit99}, the notion of accepted conclusions introduced to guarantee consistency of belief sets and its application to belief revision is new. The framework is expressive enough to represent and reason about reliability and other properties of information. It thus can be used to represent revision strategies of agents declaratively. Another advantage of the framework is that it lends itself to iteration in an obvious and natural way.

In an earlier paper \cite{Bre91} the author used nonmonotonic belief bases in the preferred subtheories framework to model revision. This approach, however, did not represent reliability information explicitly. Williams and Antoniou \cite{WilAnt98} investigated revision of Reiter default theories. In a similar spirit, Antoniou et al \cite{Antetal99} discuss revision of theories expressed in Nute's defeasible logic. Also these approaches do not reason about the reliability of information. This is also true for existing work in revising logic programs, see \cite{AlfPer96} for an example.

Forms of revision where new information is not necessarily accepted were investigated by Hansson \cite{Han97}. This form of revision is sometimes referred to as non-prioritized belief
revision. Hansson called his version "semi-revision". Explicit reasoning about the available information is not modelled in Hansson's approach.

Structured belief bases were investigated by Wassermann \cite{Was98}. Rather than using the structure to model meta-level and preference information, Wassermann uses structure to determine relevant parts of the belief base. The focus is thus on local revision operations and related complexity issues. Chopra and Parikh \cite{ChoPar99} propose a model where belief bases are partitioned into subbases according to syntactic criteria. Belnap's four-valued logic is used for query answering. Again the focus is on keeping the effects of revision as local as possible. It is assumed that the local revision operators used satisfy the AGM postulates. The approach is thus very different from ours.

\section*{Acknowledgements}
The work presented in this paper was funded by DFG (Deutsche Forschungsgemeinschaft), Forschergruppe Kommunikatives Verstehen. I thank R. Booth, S. Lange, H. Sturm and F. Wolter for helpful comments. Thanks also to the anonymous reviewers of the paper.

\bibliography{brevis}
\bibliographystyle{aaai}
\end{document}